\documentclass[acmlarge,screen,nonacm]{acmart}

\AtBeginDocument{%
  }
\usepackage{graphicx}
\usepackage{xcolor}
\usepackage{listings}
\usepackage{tcolorbox}
\usepackage{lipsum}
\usepackage{booktabs, multirow}
\usepackage{soul}
\usepackage{xcolor,colortbl} 
\usepackage{changepage,threeparttable}
\usepackage{makecell}
\usepackage{url}
\lstdefinelanguage{XML}{
  morestring=[b]",
  morestring=[s]{>}{<},
  morecomment=[s]{<?}{?>},
  stringstyle=\color{blue},
  identifierstyle=\color{red},
  keywordstyle=\color{purple},
  morekeywords={infon, key, id, document, annotation, text, location, offset, length, category},
}

\setcopyright{acmlicensed}
\copyrightyear{2025}
\acmYear{2025}
\acmDOI{10.1145/3712256.3726334}
\acmConference[GECCO '25]{Genetic and Evolutionary Computation
Conference}{July 14--18, 2025}{Malaga, Spain}
\acmISBN{}

\begin{document}

\title{FoodSEM: Large Language Model Specialized in Food Named-Entity Linking}


\author{Ana Gjorgjevikj}
\email{ana.gjorgjevikj@ijs.si}
\orcid{0000-0002-5135-7718}

\affiliation{%
  \institution{Jozef Stefan Institute}
  \city{Ljubljana}
  \country{Slovenia}
}

\author{Matej Martinc}
\email{matej.martinc@ijs.si}
\orcid{0000-0002-7384-8112}
\affiliation{%
  \institution{Jozef Stefan Institute}
  \city{Ljubljana}
  \country{Slovenia}
}

\author{Gjorgjina Cenikj}
\email{gjorgjina.cenikj@ijs.si}
\orcid{0000-0002-2723-0821}
\affiliation{%
  \institution{Jozef Stefan Institute}
  \city{Ljubljana}
  \country{Slovenia}
}
\affiliation{%
  \institution{Jozef Stefan International
Postgraduate School}
\city{Ljubljana}
\country{Slovenia}
}

\author{Sa\v{s}o D\v{z}eroski}
\email{saso.dzeroski@ijs.si}
\orcid{0000-0003-2363-712X}
\affiliation{%
  \institution{Jozef Stefan Institute}
  \city{Ljubljana}
  \country{Slovenia}
}

\author{Barbara Korou\v{s}i\'{c} Seljak}
\email{barbara.korousic@ijs.si}
\orcid{0000-0001-7597-2590}
\affiliation{%
  \institution{Jozef Stefan Institute}
  \city{Ljubljana}
  \country{Slovenia}
}

\author{Tome Eftimov}
\email{tome.eftimov@ijs.si}
\orcid{0000-0001-7330-1902}
\affiliation{%
  \institution{Jozef Stefan Institute}
  \city{Ljubljana}
  \country{Slovenia}
}

\renewcommand{\shortauthors}{Gjorgjevikj et al.}

\begin{abstract}
This paper introduces FoodSEM, a state-of-the-art fine-tuned open-source large language model (LLM) for named-entity linking (NEL) to food-related ontologies. To the best of our knowledge, food NEL is a task that cannot be accurately solved by state-of-the-art general-purpose (large) language models or custom domain-specific models/systems. Through an instruction-response (IR) scenario, FoodSEM links food-related entities mentioned in a text to several ontologies, including FoodOn, SNOMED-CT, and the Hansard taxonomy. The FoodSEM model achieves state-of-the-art performance compared to related models/systems, with F1 scores even reaching 98\% on some ontologies and datasets. The presented comparative analyses against zero-shot, one-shot, and few-shot LLM prompting baselines further highlight FoodSEM's superior performance over its non-fine-tuned version. By making FoodSEM and its related resources publicly available, the main contributions of this article include (1) publishing a food-annotated corpora into IR format suitable for LLM fine-tuning/evaluation, (2) publishing a robust model to advance the semantic understanding of text in the food domain, and (3) providing a strong baseline on food NEL for future benchmarking.
\end{abstract}




\maketitle
\section{Introduction}
\label{sec:intro}

Linking food data to ontology entities is critical in aligning it with  Findable, Accessible, Interoperable, and Reusable data (FAIR) principles and supporting the European Union (EU) goals for data-driven innovation and sustainability. Ontologies provide standardized terms/structures that make food data discoverable and efficiently accessed from diverse platforms~\cite{guarino2009ontology}. Ontologies break down the silos between agricultural, health, and nutrition data, facilitating seamless integration in line with initiatives like the European Open Science Cloud (EOSC)~\cite{budroni2019architectures}. Only in 2025, 91 new projects under Horizon Europe Cluster 6 addressed food systems and sustainability, all requiring FAIR food data management. Many researchers struggle with data interoperability due to the time-consuming linking to ontologies and not having a technical background, highlighting the need to automate this process. Annotating food data with ontology entities ensures consistency and reusability, empowering development of interpretable and reproducible artificial intelligence (AI) applications. It aligns with EU strategies like the Green Deal and Farm to Fork Strategy, promoting sustainability, traceability, and food safety. It underpins the creation of a shared agriculture and food ``data space", enabling innovation which adheres to ethical/privacy standards. 

Recently, significant progress has been made in predictive healthcare and environmental modeling, enabled by the availability of biomedical semantic vocabularies/standards, which have supported shared tasks in natural language processing (NLP) to produce annotated corpora and advance research~\cite{stenetorp2011bionlp,nedellec2013overview,arighi2011biocreative,rastegar2018biocreative}. The BioNLP Shared Tasks (2013, 2016)~\cite{nedellec2013overview} focused on biomedical topics, e.g., gene event extraction, bacteria biotopes, and plant seed regulation. BioCreative challenges~\cite{arighi2011biocreative,rastegar2018biocreative} advanced tasks like gene mention recognition, protein-protein interactions, and chemical/drug entity recognition with machine learning (ML) methods (e.g., Conditional Random Fields, Support Vector Machines, rule-based systems). BioCreative IV~\cite{mao2014overview} focused on chemical named-entity recognition (NER), while BioCreative V focused on disease-related entity relations. Later editions explored crowdsourcing, interoperability, and application in clinics and semantic text similarity. Those resources are vital to interpret health information, along with the abundant biomedical and health-related data. 

In contrast, the food and nutrition domain lacks equivalent semantic resources, crucial in the development of domain-specific AI models. Before introducing the FoodBase~\cite{popovski2019foodbase} annotated corpora, food information extraction (IE) relied on rule-based NER methods, such as drNER~\cite{eftimov2017rule} for dietary recommendations and FoodIE~\cite{popovski2019foodie} for recipe data, focusing on distinguishing between food vs. non-food entities. With FoodBase, ML approaches emerged, including BuTTER~\cite{cenikj2020butter}, the first bidirectional LSTM for food NER, and FoodNER~\cite{stojanov2021fine}, a fine-tuned BERT model for NER and simultaneous named-entity linking (NEL) tasks (linking recognized mentions of food entities in text to their corresponding ontology entities). FoodOntoMap~\cite{popovski2019foodontomap} is a resource designed to normalize and link food concepts to various food ontologies, enhancing data integration. Initially developed with mappings to four ontologies (Hansard Corpus~\cite{abercrombie2018sentiment}, FoodOn~\cite{dooley2018foodon}, OntoFood, and SNOMED Clinical Terms (SNOMED CT)~\cite{donnelly2006snomed}), it was expanded to FoodOntoMapV2~\cite{popovski2019foodontomapv2} by adding RCD, Medical Subject Headings (MeSH)~\cite{diaz2008integrating}, Systematized Nomenclature of Medicine (SNMI), and National Drug Data File (NDDF) for broader interoperability. The public dataset includes unique food concepts and mappings in CSV and Excel, supporting applications that connect food, health, and environmental research. FoodViz~\cite{stojanov2020foodviz} is a web-based tool enabling experts to identify food entity mentions in text and link them to entities from food ontologies, using FoodOntoMap data. Exact matches are linked directly, while ambiguous cases may be linked to multiple entities due to food ontology coverage gaps. FoodKG~\cite{haussmann2019foodkg} is a semantic-driven knowledge graph for food recommendation, unifying domain-specific ontologies to support healthy eating. It integrates food and recipe data while preserving provenance, using FoodOn for entity linking. 

Recently, the so-called foundation models \cite{bommasani2021opportunities} have revolutionized NLP. The development of large language models (LLMs) has enabled task-agnostic architectures to solve downstream NLP tasks in a zero-shot fashion without the need for labeled data \cite{brown2020language}. Researchers are exploring LLMs for biomedical applications, focusing on improving clinical NER through prompt engineering and fine-tuning for biomedical NLP tasks, including NER and NEL~\cite{biana2024vaner,zhu2024comparative}. LLMs have also been applied to food NER~\cite{pitsilou2024using} with promising results, but a recent evaluation of GPT models for NEL revealed that while they can identify food entity mentions, they struggle linking them to specific food ontologies~\cite{ogrinc2024zero}.

This paper presents FoodSEM, a fine-tuned open-source LLM for food NEL. Through an instruction-response (IR) scenario, FoodSEM is able to link food-related entity mentions (single or multi-word sequences) in a text given in its instruction to entities (represented through their URIs) from three food ontologies (FoodOn, SNOMED-CT, and the Hansard taxonomy). By making all resources publicly available, this paper's contributions are (1) publishing food-annotated corpora in IR format suitable for LLM fine-tuning, (2) publishing a robust model to advance the semantic understanding of text from food domain, and (3) providing a strong baseline on food NEL to serve as a benchmark dataset. FoodSEM achieves high F1 scores, up to 98\% for some ontologies. FoodSEM's comparison with 0-shot, 1-shot, and 5-shot prompting baselines demonstrates its superior performance over its non-fine-tuned version. The code and datasets are available on GitHub\footnote{\tiny \url{https://github.com/matejMartinc/FoodSEM}}. FoodSEM is available in Hugging Face's model repository\footnote{\tiny \url{https://huggingface.co/Matej/FoodSEM-LLM}}.

The paper is organized as follows. Section~\ref{sec:related_work} reviews related work on food NEL and Section~\ref{sec:food_semantic} the used ontologies and corpora. Section~\ref{sec:foodSematnica} details the fine-tuning process. Section~\ref{sec:results} presents the results, and Section~\ref{sec:conclusion} the conclusion.

\section{Related work}
\label{sec:related_work}
NEL~\cite{tedeschi2021named} is critical in integrating data across diverse domains. By associating entity mentions in text with entities' unique identifiers from ontologies, NEL provides a consistent way to reference those entities, allowing information aggregation for the same entities from multiple sources, regardless of different naming conventions or synonyms. In this direction, StandFood~\cite{eftimov2017standfood} is a semi-automatic system that links food concepts to the FoodEx2~\cite{european2015food} taxonomy, considering lexical, but not semantic similarity. The development of food-annotated corpora like FoodBase~\cite{popovski2019foodbase}, CafeteriaFCD~\cite{ispirova2022cafeteriafcd}, and CafeteriaSA~\cite{cenikj2022cafeteriasa} enabled the emergence of corpus-based NEL methods. FoodNER~\cite{stojanov2021fine} is a corpus-based framework for food NER and NEL, featuring 15 models, fine-tuned from three pre-trained BERT models on five semantic resources: food vs. nonfood entities, two subsets of Hansard entities, FoodOn, and SNOMED-CT entities. Leveraging BERT's capabilities in IE, FoodNER achieves macro F1 of 93.30\%–94.31\% for food vs. nonfood classification and 73.39\%–78.96\% for semantic entity prediction. It supports five tasks, including distinction of entities at the level of food groups using entities from Hansard, FoodOn, and SNOMED-CT, establishing a new benchmark in food NER and NEL. The FoodNER models work well on recipe instruction data, but fail to generalize on scientific articles. SciFoodNER~\cite{cenikj2022scifoodner} is a scientific food NER and NEL model fine-tuned on scientific abstracts (SAs) annotated with food entities. The model identifies food mentions in text and links them to Hansard, FoodOn, and SNOMED-CT. BioBERT~\cite{lee2020biobert} achieved the best performance, with a median macro F1 of 66\% for NEL to Hansard, 43\% to FoodOn, and 58\% to SNOMED-CT. Recent 0-shot evaluations of GPT-3.5 and GPT-4 models have shown low accuracy in food NEL~\cite{ogrinc2024zero}, too low for practical use.

\section{Food semantic resources}
\label{sec:food_semantic}
This section first describes the used food ontologies, and then the text corpora annotated with ontology entities, used to create the LLM fine-tuning dataset.

\textbf{FoodOn}~\cite{dooley2018foodon} is an open-source ontology designed to comprehensively describe entities with a ``food role", spanning the entire food system from agricultural production to consumption. It offers a structured vocabulary covering food sources, products, and processes, with detailed description of food-related entities. As part of the Open Biological and Biomedical Ontology (OBO) Foundry, FoodOn integrates seamlessly with other life science ontologies, promoting interoperability. It is organized into facets, addressing subdomains like organism parts, food processing methods, and quality attributes. \textbf{SNOMED-CT}~\cite{donnelly2006snomed} is a standardized medical ontology that provides a comprehensive vocabulary of clinical concepts (diseases, treatments, anatomy, and procedures). It contains the concept of Food that can be used for food concept normalization. Widely adopted worldwide, it facilitates integration between healthcare systems through its hierarchical structure and interoperability. Maintained by SNOMED International, it supports advanced data analysis and multilingual use, making it essential for effective health information management. The \textbf{Hansard taxonomy}~\cite{abercrombie2018sentiment} is a structured collection of concepts developed in the SAMUELS project (2014–2016) to enable semantically-driven searches within its dataset. This taxonomy organizes concepts into 37 overarching semantic groups, facilitating intuitive information retrieval. Among them, the ``Food and Drink" (denoted AG) categorizes culinary/beverage terms and concepts. By structuring data semantically, the taxonomy enhances accessibility/usability of complex textual corpora for researchers/developers, supporting advanced search features and more nuanced analyses.

The \textbf{CafeteriaFCD} corpus~\cite{ispirova2022cafeteriafcd} is a curated text dataset of food consumption data (recipes) annotated with food entities from Hansard, FoodOn, and SNOMED-CT. Built on the FoodBase corpus (containing 1,000 gold-standard annotated recipes), it has been created with the NCBO Annotator for FoodOn and SNOMED-CT entities, later manually corrected by domain experts. It contains 7,429 annotations. The recipe text varies in length (58-2024 characters, 511.8 on average). Each recipe includes at least one annotation. In relation to Hansard, the recipes contain up to 37 annotations (10.7 on average). From FoodOn, they contain up to 22 annotations (7.6 on average), and from SNOMED-CT up to 17 (5.5 on average). \textbf{CafeteriaSA}~\cite{cenikj2022cafeteriasa} is a pioneering annotated text corpus of 500 SAs, annotated with food entities from multiple ontologies. It includes 6,407 annotations from Hansard, 4,299 from FoodOn, and 3,623 from SNOMED-CT. The average abstract length is 1,467.6 characters (ranging 236-2,510 characters). From Hansard, there are on average 8.4 annotations per SA (ranging 1-27), from FoodOn 4.8 annotations (1-18), and from SNOMED-CT 3.8 annotations (1-15).

\section{FoodSEM}
\label{sec:foodSematnica}

\subsection{Fine-tuning dataset creation}

\begin{lstlisting}[caption={BioC-formatted recipe from CafeteriaFCD mapped to FoodOn ontology.}]
<document>
    <id>0recipe1006</id>
    <infon key="full_text"> Mix the cream cheese, beef, olives, onion, and Worcestershire sauce together in a bowl until evenly blended. Keeping the mixture in the bowl, scrape it into a semi-ball shape. Cover, and refrigerate until firm, at least 2 hours. Place a large sheet of waxed paper on a flat surface. Sprinkle with walnuts. Roll the cheese ball in the walnuts until completely covered. Transfer the cheese ball to a serving plate, or rewrap with waxed paper and refrigerate until needed. </infon>
    <annotation id="1"> <infon key="semantic_tags">http://purl.obolibrary.org/obo/FOODON_03301889;http://purl.obolibrary.org/obo/FOODON_00001013</infon>
        <location offset="3" length="12" /> <text>CREAM CHEESE</text> </annotation>
    <annotation id="2"> <infon key="semantic_tags">http://purl.obolibrary.org/obo/FOODON_03301704;http://purl.obolibrary.org/obo/NCBITaxon_4679</infon>
        <location offset="10" length="5" /> <text>ONION</text> </annotation>
    <annotation id="3"> <infon key="semantic_tags">http://purl.obolibrary.org/obo/FOODON_03305003;http://purl.obolibrary.org/obo/FOODON_03311146</infon>
        <location offset="13" length="20" /> <text>WORCESTERSHIRE SAUCE</text> </annotation>
    <annotation id="4"> <infon key="semantic_tags">http://purl.obolibrary.org/obo/NCBITaxon_16718</infon>
        <location offset="63" length="7" /> <text>WALNUTS</text> </annotation>
    <annotation id="5"> <infon key="semantic_tags">http://purl.obolibrary.org/obo/FOODON_00001013</infon>
        <location offset="67" length="11" /> <text>CHEESE</text> </annotation>
    <annotation id="6"> <infon key="semantic_tags">http://purl.obolibrary.org/obo/NCBITaxon_16718</infon>
        <location offset="71" length="7" /> <text>WALNUTS</text> </annotation>
    <annotation id="7"> <infon key="semantic_tags">http://purl.obolibrary.org/obo/FOODON_00001013</infon>
        <location offset="78" length="11" /> <text>CHEESE</text> </annotation>
    <infon key="category">Appetizers and snacks</infon>
</document>
\end{lstlisting}

The fine-tuning dataset is created by processing the two Cafeteria corpora in BioC XML format~\cite{comeau2013bioc}, enabling interoperability of biomedical data and annotations. The process starts by converting BioC-formatted data into a format suitable for LLM fine-tuning (converting each BioC-formatted instance into an IR pair). Listing 1.1 shows a BioC-formatted instance from CafeteriaFCD, giving the recipe text and food annotations. The annotations consist of mentions of food entities in the recipe, their location in the text, the actual mention text, and the Uniform Resource Identifiers (URIs) of the linked entities from FoodOn. The mention location in the text is described through its start offset and the number of characters. One recipe has two additional BioC-formatted instances, with all food annotations linked to Hansard and SNOMED-CT entities, respectively. Each BioC instance is converted into an IR pair sequence, including two types of instruction, asking the LLM to either: (1) identify all mentions of food entities in the provided text (which can either be a recipe or a SA), pertaining to the NER task, or (2) link all identified food entity mentions to entities (represented through URIs) from the Hansard taxonomy, FoodOn, or SNOMED-CT, pertaining to the NEL task. The first instruction is concatenated with the actual recipe/SA text in which the entity mentions should be identified, followed by the expected LLM response. The second instruction is repeated three times, each time asking the LLM to link the previously identified entity mentions with entities from one of the three ontologies. Each instruction is followed by the expected LLM response, which contains a generic starting response phrase, followed by a list of entity mentions with their linked ontology entities (represented with URIs). The described transformation of the previously shown BioC-formatted recipe is given in the box below. A sequence of four IR pairs is given, with the instruction given in tags [INST] ... [/INST]. The first pair pertains to the NER task, while the remaining three pertain to the NEL task. There are no duplicate recipe and SA instances in the fine-tuning dataset to prevent data leakage.

\begin{tcolorbox}[
  colframe=black,    
  colback=white,     
  coltitle=black,    
  title=Food Entity Linking Example,
  fonttitle=\bfseries, 
  sharp corners,     
  boxrule=0.5pt,     
  width=\textwidth,   
  fontupper=\tiny
]

\textbf{[INST]} Find all food entities present in the text:  
Mix the cream cheese, beef, olives, onion, and Worcestershire sauce together in a bowl until evenly blended. 
Keeping the mixture in the bowl, scrape it into a semi-ball shape. Cover, and refrigerate until firm, 
at least 2 hours. Place a large sheet of waxed paper on a flat surface. Sprinkle with walnuts. 
Roll the cheese ball in the walnuts until completely covered. Transfer the cheese ball to a serving plate, 
or rewrap with waxed paper and refrigerate until needed. \textbf{[/INST]} Certainly, the entities connected with food are outlined as follows:  
cream cheese, beef, olives, onion, Worcestershire sauce, walnuts, cheese ball.  

\textbf{[INST]} Is it within your capabilities to link the extracted food entities to a Hansard taxonomy? \textbf{[/INST]} Certainly, the entities are associated properly:  
cream cheese - AG.01.e [Dairy produce]; AG.01.e.02 [Cheese]; AG.01.n [Dishes and prepared food];  
AG.01.n.18 [Preserve],  
beef - AG.01.d.03 [Beef],  
olives - AG.01.h.01.e [Fruit containing stone],  
onion - AG.01.h.02.e [Onion/leek/garlic],  
Worcestershire sauce - AG.01.h [Fruit and vegetables]; AG.01.l.04 [Sauce/dressing],  
walnuts - AG.01.h.01.f [Nut],  
cheese ball - AG.01.e.02 [Cheese]; AG.01.n.18 [Preserve].  

\textbf{[INST]} Might I trouble you to connect the extracted food entities to a FoodOn ontology, if possible? \textbf{[/INST]} Definitely, the entities are linked suitably:  
cream cheese - FOODON-03301889; FOODON-00001013,  
onion - FOODON-03301704;  
NCBITaxon-4679,  
worcestershire sauce - FOODON-03305003;  
FOODON-03311146,  
walnuts - NCBITaxon-16718,  
cheese - FOODON-00001013.  

\textbf{[INST]} Might it be possible for you to link them to a SNOMEDCT ontology, if needed? \textbf{[/INST]} Absolutely, the entities are related as such:  
cream cheese - SNOMEDCT-226849005;  
SNOMEDCT-255621006; SNOMEDCT-102264005,  
beef - SNOMEDCT-226916002,  
olives - SNOMEDCT-227436000,  
onion - SNOMEDCT-735047000,  
worcestershire sauce - SNOMEDCT-443701000124100;  
SNOMEDCT-227519005,  
cheese - SNOMEDCT-102264005.  

\end{tcolorbox}

\begin{figure}[!t]
    \centering
    \includegraphics[trim={0 0 0 2cm}, clip, width=0.7\linewidth]{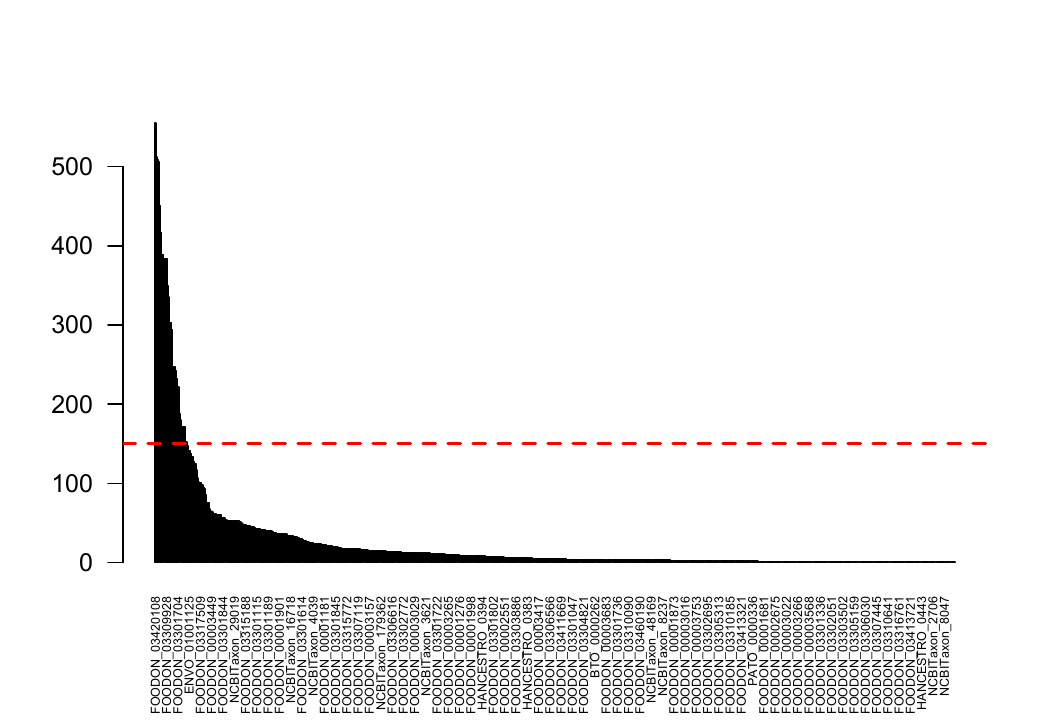}
    \caption{Distribution of FoodON entities in CafeteriaFCD. Only a subset of entity names is displayed on the x-axis. The red line represents the threshold of 150.}
    \label{fig:distibution_FoodON}
    \vspace{-0.2cm}
\end{figure}

To create diverse instruction phrases that precede the recipe or SA text, a pool of 1,000 instruction phrase variations was generated with GPT-3.5. An instruction phrase was randomly selected for each BioC-formatted instance from the corpora and concatenated with the recipe/SA text. Each response that follows the instruction also begins with a generic phrase, followed by a list of food entity mentions and their linked entity URIs from the specified ontology. The beginning response phrase was also randomly selected from a separate pool of 1,000 GPT-3.5-generated phrase variations and concatenated with the linked entity URIs. After converting the BioC-formatted instances into IR pair sequences, we analyzed the entity coverage across the ontologies in the fine-tuning datasets. This analysis was performed only on CafeteriaFCD, as it contains recipe data with a greater diversity of food entities compared to the SAs from CafeteriaSA. The analysis showed underrepresented entities, for which artificially generated IR instances were added to the fine-tuning dataset for a balanced distribution. Figure~\ref{fig:distibution_FoodON} illustrates the FoodON entity distribution in CafeteriaFCD. Similar analysis was done for the other two ontologies, but is not shown due to space limitations. Based on the distributions, we set a desired threshold of 150 mentions by entity in the fine-tuning dataset (to be refined in the future through sensitivity analysis). If an entity has fewer than 150 mentions ($k<150$), we include it in the generation of artificial IR pairs until it reaches the threshold. Unlike the IR pair sequences generated from CafeteriaFCD/SA, which involve one NER instruction and three NEL instructions, the artificially generated IR pairs do not include a NER instruction, only one NEL instruction. To achieve this, we generated a sample of food entity labels for each entity within each ontology, with a size of $150-k$. Next, we combined all labels across entities within each ontology, shuffled them, and created sets with 7, 9, and 12 labels chosen randomly. Each set was used to generate three IR pairs, each instructing the LLM to link the labels to one of the three ontologies. We generate 13,492 artificial IR pairs for NEL to FoodOn, 1,611 to Hansard, and 4,445 to SNOMED-CT, making sure that no duplicate instances are generated, to avoid data leakage during LLM cross-validation (CV) (see Section \ref{sec:experimental-design}). We note again that CafeteriaFCD/SA IR sequences include NER and NEL IR pairs, while the artificially generated sequences include only NEL IR pairs. The final fine-tuning dataset consists of 1,479 Cafeteria IR sequences, 13,492 artificial FoodOn NEL IR pairs, 1,611 artificial Hansard NEL IR pairs, 4,445 artificial SNOMED-CT NEL IR pairs, or a total of 21,027 fine-tuning instances. FoodSEM is expected to correctly do NEL to three ontologies, so it is crucial that it sees all entity URIs from those ontologies during fine-tuning. In such a way, it learns to link single/multi-word mentions to correct entity URIs without hallucinations. The model is not designed to find links to entities absent from the three ontologies, as NEL is an information retrieval task.

\subsection{Fine-tuning process}

We use the instruction-tuned 8B parameter version of the Meta Llama 3 family of LLMs~\cite{dubey2024llama}. Llama 3 is an auto-regressive language model using an optimized transformer architecture \cite{vaswani2017attention}. The model was pre-trained on a dataset with over 15 trillion tokens, sourced from publicly available content and curated using advanced filtering methods like heuristic filters and semantic deduplication. The 8B model has a knowledge cutoff of March 2023. The instruction-tuned version we use\footnote{\tiny We use the model available on HuggingFace: \url{https://huggingface.co/meta-llama/Meta-Llama-3-8B-Instruct}} is optimized for dialogue use cases by applying an alignment process including supervised fine-tuning, rejection sampling, proximal policy optimization, and direct preference optimization. It was instruction-tuned on over 10 million manually labeled data points, curated to improve reasoning, code generation, and similar. We selected an open-source LLM since proprietary LLMs (e.g., GPT-4, Gemini, or Claude) cannot be fine-tuned for specific tasks/domains. With our limited GPU resources, it was infeasible to opt for a bigger model (e.g., 70B or 400B parameter model), as it would significantly increase training time/cost. We chose the Llama 3 8B model due to its competitive performance on open-source LLM benchmarks\footnote{\tiny \url{https://huggingface.co/spaces/open-llm-leaderboard/open_llm_leaderboard}} when conducting our experiments. To reduce memory usage and speed up the fine-tuning process, we load the model in 4-bit precision and fine-tune it using Low-Rank Adaptation (LoRA)~\cite{hu2021lora}. We freeze all layers in the backbone model and train only low-rank perturbations to query and value weight matrices in the model. We train the model using a supervised fine-tuning (SFT) trainer, using the next word prediction objective. The fine-tuning hyperparameters are: learning rate of 2e-4, LoRA intrinsic rank of 16, LoRA alpha of 16, LoRA dropout of 0.05, and sequence length of 1024. We fine-tuned the model for one epoch using batch size of 10, and 10 warm-up steps. The input sequences were padded to the longest sequence in the batch and truncated if longer than 1024 tokens. By default, the Llama tokenizer does not have a padding token, therefore we added it by using one of the designated extra token spots, reserved for additional special tokens (we use the ``<|reserved\_special\_token\_250|>'' spot\footnote{\tiny Initial experiments showed that a popular practice of setting the ``end of sequence'' token as a padding token resulted in model generating repetitions and having problems with finishing the generated sequence.}).

\section{Evaluation}
\label{sec:results}

\subsection{Experimental design}
\label{sec:experimental-design}

We evaluate FoodSEM's performance in a CV setting. As per Section~\ref{sec:foodSematnica}, the fine-tuning uses four IR pair instance datasets: (1) NEL to Hansard taxonomy (consisting of an instruction to link food entity mentions to entities from Hansard taxonomy and expected response), (2) NEL to FoodON, (3) NEL to SNOMED-CT, and (4) NER of mentions of food entities in recipes/SAs. Each of the datasets is divided in five random chunks with the Scikit-learn library. For each CV fold, we construct a dataset-specific training set by concatenating 4 training chunks and using the remaining fifth for testing. The final training set per CV fold is constructed by concatenating (and shuffling) 4 datasets-specific training sets. The test set for each fold is constructed by concatenating all datasets-specific test folds. Therefore, the training set in each fold consists of 80\% of the data and the test from the remaining 20\%. We do not use CV folds for hyperparameter search or early stopping. To avoid catastrophic forgetting \cite{luo2023empirical} and prevent overfitting, we add in each CV fold's training set IR instances from the Dolly-HHRLHF subset \cite{mosaicml2023dolly_hhrlhf} of the open source MosaicML Instruct-v3 dataset \cite{mosaicml2023instruct-v3}. The Dolly-HHRLHF dataset is a general instruction-tuning dataset generated by Databricks\footnote{\tiny \url{https://www.databricks.com}} employees, covering different tasks from domains, such as brainstorming, open/closed question-answering, and summarization. Specifically, into the training set of each CV fold, we randomly shuffle in 34,229 IR instances from Dolly-HHRLHF, filtered according to length (all instances shorter or equal to the max sequence length of 1024 tokens). Using the five training sets, we obtain five fine-tuned models, tested on the appropriate test set. During testing, the model is given an instruction part of the IR instance as input, and the generated response is compared to the expected one. Note that for both Cafeteria corpora, the NEL task follows a NER task, so the model is, in the first step, instructed to extract food entity mentions. In the second step, the extracted mentions are fed to the model as input, with an instruction to link them to an ontology (Hansard, FoodON, or SNOMED-CT). Fine-tuning one model instance took $\sim$48 hours on an NVIDIA A100 80GB GPU. The entire fine-tuning of five models took about 240 hours of GPU time.

The fine-tuned models were compared to the non-fine-tuned on each of the five test folds. Scenarios involving 0-shot, 1-shot, and 5-shot prompting of the non-fine-tuned model were used as baselines. In each, the model was prompted with an instruction part of the test IR instance concatenated with none, one, or five example IR instances sampled randomly from the matching training set, respectively. The IR instance(s) were selected with regard to Hansard, FoodOn, SNOMED-CT, or food NER, depending on the given input test set instruction. An example of a 1-shot instruction prompt is shown in the box below.

\begin{tcolorbox}[
  colframe=black,    
  colback=white,     
  coltitle=black,    
  title=Food Entity Linking Example,
  fonttitle=\bfseries, 
  sharp corners,     
  boxrule=0.5pt,     
  width=\textwidth,   
  fontupper=\tiny
]

\textbf{[INST]} The following are examples of questions (with answers) about nutrition. \textbf{Question}: I kindly ask for links to the FoodOn ontology for the mentioned entities: cheddar cheese, cookie dough, flax?
\textbf{Answer}: Absolutely, the entities are related properly: cheddar cheese - http://purl.obolibrary.org/obo/FOODON\_03302458; http://purl.obolibrary.org/obo/FOODON\_00001013, cookie dough - http://purl.obolibrary.org/obo/FOODON\_03310689; http://purl.obolibrary.org/obo/FOODON\_00002466; http://purl.obolibrary.org/obo/FOODON\_03301585; http://purl.obolibrary.org/obo/FOODON\_03311552, flax - http://purl.obolibrary.org/obo/NCBITaxon\_4006. \\
Respond to the following question in the same manner as seen in the examples above. \textbf{Question}: Could you provide links to the FoodOn ontology for these entities: green tea, oil, salmon, flax, lettuce, eggnog, spring, flaxseed meal, maraschino cherry? 
\textbf{Answer}:  \textbf{[/INST]}
\end{tcolorbox}

Custom LLM response parsers were developed to extract predictions (entity URIs) even when responses do not follow the formatting the LLM was fine-tuned to output. This also applies to the baseline 0-shot, 1-shot, and 5-shot responses of the non-fine-tuned LLM. Only entity mentions for which a URI is given in the true response are taken in consideration during the evaluation. Incorrect entity URI linking, non-meaningful responses (e.g., empty ones, hallucinations that contain no entity URIs for the mentions in the instruction) were counted as errors. 

\subsection{Results and Discussion}
The LLM performance was evaluated with three metrics: macro-weighted NEL precision, macro-weighted NEL recall, and macro-weighted NEL F1. In NEL, precision measures the proportion of correctly linked entities out of all entities linked by the LLM, i.e., its ability to avoid false positives. Recall evaluates the proportion of correctly linked entities out of all ground truth entities, i.e., the ability to link all entities mentioned in the text. The F1 combines precision and recall in a single metric, offering a holistic measure of the LLM's linking accuracy. Although we tried to balance entity distribution across each ontology through artificial IR instances, the dataset is imbalanced overall, as some entities appear more than 150 times in the Cafeteria corpora and were not removed from the dataset. To alleviate this issue, we calculate the macro-weighted average of the metrics, derived by taking the weighted mean of precision, recall, or F1 for each entity, factoring in the actual frequency of each entity in the test dataset.

Table~\ref{tab:FoodSem_results} gives FoodSEM's performance on artificially generated, CafeteriaFCD, and CafeteriaSA IR instances. As the artificial IR instances concern the NEL task only, and for both Cafeteria corpora, the NEL task is preceded by an NER task, any entity mentions not recognized during the NER task are excluded from the NEL process, which then affects the NEL performance. For each test dataset (learning task + ontology), results are presented based on five-fold CV. On the \textbf{artificially generated} IR instances, FoodSEM has consistently high and robust performance across all metrics. On the artificial instances, on FoodOn the average macro-weighted precision is 0.953($\pm$0.002), recall 0.936($\pm$0.002), and F1 0.942($\pm$0.002). For SNOMED-CT, they are 0.981($\pm$0.003), 0.969($\pm$0.003), and 0.974($\pm$0.002), and for Hansard taxonomy 0.948($\pm$0.004), 0.929($\pm$0.011), and 0.937($\pm$0.007), accordingly. On \textbf{CafeteriaFCD}, FoodSEM results for FoodOn are 0.923($\pm$0.004), 0.770($\pm$0.018), and 0.827($\pm$0.012), for SNOMED-CT 0.956 ($\pm$0.010), 0.804($\pm$0.024), and 0.863($\pm$0.019), and for Hansard taxonomy, 0.915 ($\pm$0.013), 0.791($\pm$0.010), and F1 0.843($\pm$0.010), accordingly. On \textbf{CafeteriaSA}, FoodSEM results for FoodOn are 0.804($\pm$0.029), 0.580($\pm$0.035), and 0.654($\pm$0.029), for SNOMED-CT 0.849($\pm$0.019), 0.640($\pm$0.025), and 0.713($\pm$0.022), for Hansard 0.872($\pm$0.016), 0.653($\pm$0.035), and 0.730($\pm$0.026), accordingly. From the results for CafeteriaFCD and CafeteriaSA, it can be concluded that the precision for each ontology is consistently high and robust, although it slightly decreases on CafeteriaSA. This suggests that most of the entities linked by FoodSEM are relevant. However, the recall is somewhat lower, indicating that not all entity mentions in the SA are successfully linked. This is likely due to errors in the NER task, which are propagated to the NEL task, i.e., if an entity mention is not extracted in the NER task, the NEL task cannot link it, affecting NEL F1.

\begin{table}[t!]
\centering
\caption{FoodSEM results on artificial, CafeteriaFCD, and CafeteriaSA IR instances.}\label{tab:FoodSem_results}
\scriptsize
\resizebox{\textwidth}{!}{
\begin{tabular}{llrrrllrrrllrrr}\toprule
\multicolumn{5}{l}{\makecell[l]{Artificial NEL Samples}} & &
\multicolumn{4}{l}{\makecell[l]{CafeteriaFCD}} & &
\multicolumn{4}{l}{\makecell[l]{CafeteriaSA}} \\\cmidrule{1-15}
\makecell[l]{Test\\ Fold} &
\makecell[l]{Test\\ Set} &
\makecell[r]{Macro\\ Weighted\\ Precision} &
\makecell[r]{Macro\\ Weighted\\ Recall} &
\makecell[r]{Macro\\ Weighted\\ F1} & &
\makecell[l]{Test\\ Set} &
\makecell[r]{Macro\\ Weighted\\ Precision} &
\makecell[r]{Macro\\ Weighted\\ Recall} &
\makecell[r]{Macro\\ Weighted\\ F1} & &
\makecell[l]{Test\\ Set} &
\makecell[r]{Macro\\ Weighted\\ Precision} &
\makecell[r]{Macro\\ Weighted\\ Recall} &
\makecell[r]{Macro\\ Weighted\\ F1} \\\cmidrule{1-15}
1 & \multirow{6}{*}{\makecell[l]{NEL\\ (FoodOn)}} & 
0.954 & 0.934 & 0.941 & &
\multirow{6}{*}{\makecell[l]{NER+NEL\\ (FoodOn)}} & 
0.923 & 0.769 & 0.823 & &
\multirow{6}{*}{\makecell[l]{NER+NEL\\ (FoodOn)}} & 
0.787 & 0.523 & 0.609 \\
2 & & 0.953 & 0.933 & 0.940 & &
& 0.928 & 0.746 & 0.811 & &
& 0.825 & 0.577 & 0.657 \\
3 & & 0.951 & 0.937 & 0.942 & &
& 0.927 & 0.759 & 0.824 & &
& 0.812 & 0.608 & 0.675 \\
4 & & 0.955 & 0.938 & 0.944 & &
& 0.917 & 0.786 & 0.836 & &
& 0.762 & 0.584 & 0.645 \\
5 & & 0.952 & 0.937 & 0.943 & &
& 0.921 & 0.790 & 0.840 & &
& 0.833 & 0.608 & 0.683 \\\cmidrule{1-15}
Avg & & 0.953(0.002) & 0.936(0.002) & 0.942(0.002) & &
& 0.923(0.004) & 0.770(0.018) & 0.827(0.012) & &
& 0.804(0.029) & 0.580(0.035) & 0.654(0.029) \\
\bottomrule
1 & \multirow{6}{*}{\makecell[l]{NEL\\ (Snomed)}} & 
0.979 & 0.966 & 0.972 & &
\multirow{6}{*}{\makecell[l]{NER+NEL\\ (Snomed)}} & 
0.969 & 0.826 & 0.882 & &
\multirow{6}{*}{\makecell[l]{NER+NEL\\ (Snomed)}} & 
0.827 & 0.612 & 0.685 \\
2 & & 0.978 & 0.971 & 0.974 & &
& 0.950 & 0.782 & 0.847 & &
& 0.870 & 0.660 & 0.735 \\
3 & & 0.981 & 0.967 & 0.973 & &
& 0.944 & 0.774 & 0.840 & &
& 0.857 & 0.670 & 0.736 \\
4 & & 0.984 & 0.970 & 0.976 & &
& 0.955 & 0.818 & 0.872 & &
& 0.831 & 0.636 & 0.703 \\
5 & & 0.983 & 0.972 & 0.977 & &
& 0.960 & 0.821 & 0.876 & &
& 0.858 & 0.621 & 0.707 \\\cmidrule{1-15}
Avg & & 0.981(0.003) & 0.969(0.003) & 0.974(0.002) & &
& 0.956(0.010) & 0.804(0.024) & 0.863(0.019) & &
& 0.849(0.019) & 0.640(0.025) & 0.713(0.022) \\
\bottomrule
1 & \multirow{6}{*}{\makecell[l]{NEL\\ (Hansard)}} & 
0.950 & 0.935 & 0.941 & &
\multirow{6}{*}{\makecell[l]{NER+NEL\\ (Hansard)}} & 
0.922 & 0.802 & 0.854 & &
\multirow{6}{*}{\makecell[l]{NER+NEL\\ (Hansard)}} & 
0.851 & 0.594 & 0.685 \\
2 & & 0.945 & 0.909 & 0.925 & &
& 0.911 & 0.777 & 0.834 & &
& 0.865 & 0.665 & 0.731 \\
3 & & 0.944 & 0.934 & 0.938 & &
& 0.929 & 0.800 & 0.854 & &
& 0.878 & 0.688 & 0.754 \\
4 & & 0.954 & 0.935 & 0.942 & &
& 0.917 & 0.786 & 0.842 & &
& 0.895 & 0.654 & 0.743 \\
5 & & 0.947 & 0.933 & 0.938 & &
& 0.895 & 0.788 & 0.833 & &
& 0.871 & 0.665 & 0.736 \\\cmidrule{1-15}
Avg & & 0.948(0.004) & 0.929(0.011) & 0.937(0.007) & &
& 0.915(0.013) & 0.791(0.010) & 0.843(0.010) & &
& 0.872(0.016) & 0.653(0.035) & 0.73(0.026) \\
\bottomrule
\end{tabular}
}
\end{table}

\begin{table}[t!]
\centering
\caption{Non-fine-tuned LLM results in 5-shot prompting baseline scenario.}\label{tab:five-shot}
\scriptsize
\resizebox{\textwidth}{!}{
\begin{tabular}{llrrrllrrrllrrr}\toprule
\multicolumn{5}{l}{\makecell[l]{Artificial NEL Samples}} & &
\multicolumn{4}{l}{\makecell[l]{CafeteriaFCD}} & &
\multicolumn{4}{l}{\makecell[l]{CafeteriaSA}} \\\cmidrule{1-15}
\makecell[l]{Test\\ Fold} &
\makecell[l]{Test\\ Set} &
\makecell[r]{Macro\\ Weighted\\ Precision} &
\makecell[r]{Macro\\ Weighted\\ Recall} &
\makecell[r]{Macro\\ Weighted\\ F1} & &
\makecell[l]{Test\\ Set} &
\makecell[r]{Macro\\ Weighted\\ Precision} &
\makecell[r]{Macro\\ Weighted\\ Recall} &
\makecell[r]{Macro\\ Weighted\\ F1} & &
\makecell[l]{Test\\ Set} &
\makecell[r]{Macro\\ Weighted\\ Precision} &
\makecell[r]{Macro\\ Weighted\\ Recall} &
\makecell[r]{Macro\\ Weighted\\ F1} \\\cmidrule{1-15}
1 & \multirow{6}{*}{\makecell[l]{NEL\\(FoodOn)}} & 
0.368 & 0.055 & 0.093 & &
\multirow{6}{*}{\makecell[l]{NER+NEL\\(FoodOn)}} & 
0.515 & 0.249 & 0.320 & &
\multirow{6}{*}{\makecell[l]{NER+NEL\\(FoodOn)}} & 
0.356 & 0.113 & 0.160 \\
2 & & 0.367 & 0.064 & 0.106 & &
& 0.504 & 0.248 & 0.318 & &
& 0.382 & 0.113 & 0.165 \\
3 & & 0.365 & 0.061 & 0.103 & &
& 0.541 & 0.246 & 0.322 & &
& 0.326 & 0.106 & 0.149 \\
4 & & 0.352 & 0.056 & 0.095 & &
& 0.527 & 0.255 & 0.326 & &
& 0.351 & 0.139 & 0.191 \\
5 & & 0.358 & 0.057 & 0.096 & &
& 0.540 & 0.247 & 0.324 & &
& 0.319 & 0.101 & 0.142 \\\cmidrule{1-15}
Avg & & 0.362(0.007) & 0.059(0.004) & 0.099(0.006) & &
& 0.525(0.016) & 0.249(0.004) & 0.322(0.003) & &
& 0.347(0.025) & 0.114(0.015) & 0.161(0.019) \\
\bottomrule
1 & \multirow{6}{*}{\makecell[l]{NEL\\(Snomed)}} & 
0.421 & 0.127 & 0.189 & &
\multirow{6}{*}{\makecell[l]{NER+NEL\\(Snomed)}} & 
0.510 & 0.242 & 0.314 & &
\multirow{6}{*}{\makecell[l]{NER+NEL\\(Snomed)}} & 
0.442 & 0.194 & 0.256 \\
2 & & 0.432 & 0.134 & 0.196 & &
& 0.490 & 0.249 & 0.317 & &
& 0.411 & 0.164 & 0.224 \\
3 & & 0.407 & 0.126 & 0.185 & &
& 0.486 & 0.231 & 0.301 & &
& 0.388 & 0.179 & 0.235 \\
4 & & 0.400 & 0.122 & 0.181 & &
& 0.461 & 0.236 & 0.299 & &
& 0.331 & 0.144 & 0.187 \\
5 & & 0.418 & 0.127 & 0.187 & &
& 0.473 & 0.219 & 0.291 & &
& 0.386 & 0.153 & 0.208 \\\cmidrule{1-15}
Avg & & 0.416(0.012) & 0.127(0.004) & 0.188(0.006) & &
& 0.484(0.018) & 0.235(0.011) & 0.304(0.011) & &
& 0.392(0.041) & 0.167(0.020) & 0.222(0.026) \\
\bottomrule
1 & \multirow{6}{*}{\makecell[l]{NEL\\(Hansard)}} & 
0.396 & 0.069 & 0.114 & &
\multirow{6}{*}{\makecell[l]{NER+NEL\\(Hansard)}} & 
0.530 & 0.377 & 0.425 & &
\multirow{6}{*}{\makecell[l]{NER+NEL\\(Hansard)}} & 
0.377 & 0.232 & 0.275 \\
2 & & 0.383 & 0.072 & 0.116 & &
& 0.491 & 0.354 & 0.395 & &
& 0.482 & 0.264 & 0.314 \\
3 & & 0.421 & 0.074 & 0.123 & &
& 0.512 & 0.357 & 0.406 & &
& 0.487 & 0.265 & 0.331 \\
4 & & 0.417 & 0.074 & 0.121 & &
& 0.545 & 0.387 & 0.434 & &
& 0.533 & 0.291 & 0.344 \\
5 & & 0.405 & 0.067 & 0.111 & &
& 0.534 & 0.372 & 0.427 & &
& 0.492 & 0.293 & 0.341 \\\cmidrule{1-15}
Avg & & 0.404(0.016) & 0.071(0.003) & 0.117(0.005) & &
& 0.522(0.021) & 0.369(0.014) & 0.417(0.016) & &
& 0.474(0.058) & 0.269(0.025) & 0.321(0.028) \\
\bottomrule
\end{tabular}
}
\end{table}

\begin{table}[t!]
\centering
\caption{Non-fine-tuned LLM performance in 1-shot prompting baseline scenario.}\label{tab:one-shot}
\scriptsize
\resizebox{\textwidth}{!}{
\begin{tabular}{llrrrllrrrllrrr}\toprule
\multicolumn{5}{l}{\makecell[l]{Artificial NEL Samples}} & &
\multicolumn{4}{l}{\makecell[l]{CafeteriaFCD}} & &
\multicolumn{4}{l}{\makecell[l]{CafeteriaSA}} \\\cmidrule{1-15}
\makecell[l]{Test\\ Fold} &
\makecell[l]{Test\\ Set} &
\makecell[r]{Macro\\ Weighted\\ Precision} &
\makecell[r]{Macro\\ Weighted\\ Recall} &
\makecell[r]{Macro\\ Weighted\\ F1} & &
\makecell[l]{Test\\ Set} &
\makecell[r]{Macro\\ Weighted\\ Precision} &
\makecell[r]{Macro\\ Weighted\\ Recall} &
\makecell[r]{Macro\\ Weighted\\ F1} & &
\makecell[l]{Test\\ Set} &
\makecell[r]{Macro\\ Weighted\\ Precision} &
\makecell[r]{Macro\\ Weighted\\ Recall} &
\makecell[r]{Macro\\ Weighted\\ F1} \\\midrule
1 & \multirow{6}{*}{\makecell[l]{NEL\\(FoodOn)}} & 
0.195 & 0.034 & 0.055 & &
\multirow{6}{*}{\makecell[l]{NER+NEL\\(FoodOn)}} & 
0.347 & 0.067 & 0.107 & &
\multirow{6}{*}{\makecell[l]{NER+NEL\\(FoodOn)}} & 
0.236 & 0.039 & 0.062 \\
2 & & 0.200 & 0.036 & 0.059 & &
& 0.335 & 0.080 & 0.124 & &
& 0.181 & 0.021 & 0.034 \\
3 & & 0.202 & 0.032 & 0.054 & &
& 0.344 & 0.083 & 0.130 & &
& 0.18 & 0.028 & 0.046 \\
4 & & 0.208 & 0.036 & 0.059 & &
& 0.293 & 0.070 & 0.109 & &
& 0.171 & 0.022 & 0.038 \\
5 & & 0.189 & 0.035 & 0.057 & &
& 0.326 & 0.061 & 0.096 & &
& 0.190 & 0.029 & 0.047 \\\cmidrule{1-15}
Avg & & 0.199(0.007) & 0.035(0.002) & 0.057(0.002) & &
& 0.329(0.022) & 0.072(0.009) & 0.113(0.014) & &
& 0.192(0.026) & 0.028(0.007) & 0.045(0.011) \\
\bottomrule
1 & \multirow{6}{*}{\makecell[l]{NEL\\(Snomed)}} & 
0.183 & 0.045 & 0.069 & &
\multirow{6}{*}{\makecell[l]{NER+NEL\\(Snomed)}} & 
0.277 & 0.061 & 0.097 & &
\multirow{6}{*}{\makecell[l]{NER+NEL\\(Snomed)}} & 
0.229 & 0.035 & 0.055 \\
2 & & 0.208 & 0.048 & 0.075 & &
& 0.254 & 0.057 & 0.091 & &
& 0.177 & 0.026 & 0.044 \\
3 & & 0.220 & 0.055 & 0.084 & &
& 0.279 & 0.059 & 0.093 & &
& 0.210 & 0.036 & 0.061 \\
4 & & 0.216 & 0.056 & 0.085 & &
& 0.286 & 0.064 & 0.101 & &
& 0.211 & 0.031 & 0.051 \\
5 & & 0.217 & 0.057 & 0.086 & &
& 0.286 & 0.062 & 0.098 & &
& 0.235 & 0.043 & 0.070 \\\cmidrule{1-15}
Avg & & 0.209(0.015) & 0.052(0.005) & 0.080(0.007) & &
& 0.276(0.013) & 0.061(0.003) & 0.096(0.004) & &
& 0.212(0.023) & 0.034(0.006) & 0.056(0.010) \\
\bottomrule
1 & \multirow{6}{*}{\makecell[l]{NEL\\(Hansard)}} & 
0.175 & 0.114 & 0.130 & &
\multirow{6}{*}{\makecell[l]{NER+NEL\\(Hansard)}} & 
0.304 & 0.131 & 0.177 & &
\multirow{6}{*}{\makecell[l]{NER+NEL\\(Hansard)}} & 
0.206 & 0.076 & 0.105 \\
2 & & 0.141 & 0.102 & 0.112 & &
& 0.311 & 0.128 & 0.173 & &
& 0.282 & 0.071 & 0.104 \\
3 & & 0.163 & 0.100 & 0.116 & &
& 0.338 & 0.161 & 0.210 & &
& 0.211 & 0.068 & 0.098 \\
4 & & 0.147 & 0.104 & 0.116 & &
& 0.313 & 0.158 & 0.199 & &
& 0.357 & 0.094 & 0.138 \\
5 & & 0.169 & 0.116 & 0.131 & &
& 0.304 & 0.130 & 0.175 & &
& 0.289 & 0.077 & 0.111 \\\cmidrule{1-15}
Avg & & 0.159(0.014) & 0.107(0.007) & 0.121(0.009) & &
& 0.314(0.014) & 0.142(0.016) & 0.187(0.017) & &
& 0.269(0.063) & 0.077(0.010) & 0.111(0.016) \\
\bottomrule
\end{tabular}
}
\end{table}

\begin{figure}[!t]
    \centering
    \includegraphics[width=0.85\linewidth]{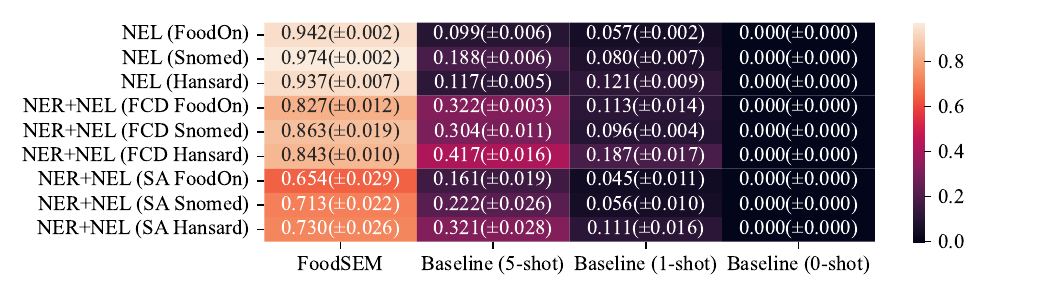}
    \caption{Heatmap of macro-weighted F1. The benchmarking scenarios are presented in rows, while FoodSEM and baseline models (0-shot, 1-shot, and 5-shot) are shown in columns.}
    \label{fig:macroF1}
    \vspace{-0.4cm}
\end{figure}

We compared FoodSEM with the non-fine-tuned model using three baseline evaluation scenarios, the same metrics, and the same folds. Results for the 5-shot prompting baseline scenario are presented in Table~\ref{tab:five-shot}, while for the 1-shot prompting scenario in Table~\ref{tab:one-shot}. The results from the 0-shot prompting were 0.0 across all metrics/ontologies, so they are not given separately. The tables highlight FoodSEM's superior performance over its non-fine-tuned version. For the 5-shot prompting baseline on the NEL task, the macro-weighted F1 scores are 0.099($\pm$0.006) for FoodOn, 0.188($\pm$0.006) for SNOMED-CT, and 0.117($\pm$0.005) for Hansard. In comparison, the 1-shot baseline yields macro-weighted F1 of 0.057($\pm$0.002) for FoodOn, 0.080($\pm$0.007) for SNOMED-CT, and 0.121($\pm$0.009) for Hansard. For CafeteriaFCD, the 5-shot baseline gives a macro-weighted F1 of 0.322($\pm$0.003) for FoodOn, 0.304($\pm$0.011) for SNOMED-CT, and 0.417($\pm$0.016) for Hansard. For CafeteriaSA, the performance of F1 is 0.161($\pm$0.019) for FoodOn, 0.222($\pm$0.026) for SNOMED-CT, and 0.321($\pm$0.028) for Hansard. In comparison, the 1-shot baseline achieves a macro-weighted F1 of 0.045($\pm$0.011) for FoodOn, 0.056($\pm$0.010) for SNOMED-CT, and 0.111($\pm$0.016) for Hansard on CafeteriaSA. On CafeteriaFCD, the F1 is 0.113($\pm$0.014) for FoodOn, 0.096($\pm$0.004) for SNOMED-CT, and 0.187($\pm$0.017) for Hansard. Overall, the 5-shot prompting baseline achieves a max F1 around 0.40, while the 1-shot baseline scores significantly lower. Figure~\ref{fig:macroF1} presents a heatmap of the mean macro-weighted F1, showing FoodSEM's superior results.

The baseline results suggest that the multi-stage reasoning (NER followed by NEL), gives the LLM additional context and leads to better results, a hypothesis to be tested in future work. The results also show that linking to Hansard tends to be easier for the non-fine-tuned model than for the other ontologies. The reason may be mostly syntactic, since linking to FoodOn and SNOMED-CT requires generating long URIs, which do not resemble standard natural language on which LLMs are pre-trained. This claim is also supported by observed examples of deterioration of the syntactic capabilities of the baseline LLM when prompts containing several long URIs (in the randomly sampled 1-shot or 5-shot IR instances) are given to the LLM. 

Here we also reference NEL results achieved on CafeteriaFCD with BERT and BioBERT~\cite{stojanov2021fine}. The macro-weighted F1 for the combined NER+NEL tasks are 0.781 for FoodOn (achieved with BERT), 0.789 for Hansard (with BioBERT-large), and 0.761 for SNOMED-CT (with BERT). The macro-weighted F1 of BioBERT on CafeteriaSA are 0.430 for FoodOn, 0.660 for Hansard, and 0.580 for SNOMED-CT~\cite{cenikj2022scifoodner}. Although the folds used in those studies differ from ours, the results indicate the expected performance. These methods have been introduced by the authors of this paper in their previous attempt of addressing NEL. The current study overcomes the challenges faced in our earlier attempts to address the NEL task. 

\section{Conclusion}
\label{sec:conclusion}
We present FoodSEM, a fine-tuned open-source LLM for food named-entity linking (NEL) that achieves state-of-the-art accuracy by linking text entities to multiple food ontologies. Supported by a dedicated instruction-response dataset, FoodSEM outperforms 0-, 1-, and 5-shot baselines, reaching F1 scores up to 98\%. Future work will extend its scope with scientific datasets and explore retrieval-augmented generation (RAG) to identify unseen entities in evolving ontologies.

\subsubsection*{Acknowledgment} 
The authors acknowledge the support of the Slovenian Research Agency through program grants No. P2-0098 and No. P2-0103, project grant No. GC-0001, as well as the Horizon Europe EU research and innovation framework programme FishEuTrust (101060712), LLMs4EU (101198470), ERA Chair AutoLearn-SI (101187010), and  MSCA-PF AutoLLMSelect (101211695).

\bibliographystyle{ACM-Reference-Format}

\appendix

\end{document}